\documentclass{llncs}

\usepackage{graphicx}
\usepackage{amsmath}
\usepackage{amssymb}
\usepackage{mathtools}
\usepackage{xcolor,colortbl}
\usepackage{booktabs}
\usepackage{listings}

\definecolor{LightGray}{gray}{0.85}
\definecolor{DarkGray}{gray}{0.65}

\newcolumntype{a}{>{\columncolor{LightGray}}c}

\begin{document}

\title{Fast and scalable learning of neuro-symbolic representations of biomedical knowledge}
\titlerunning{Scalable biomedical neuro-symbolic representations}  
%
\author{Asan Agibetov\inst{1} 
 \and Matthias Samwald\inst{1}}
\authorrunning{Asan Agibetov et al.} 
%
%
\institute{$^1$Section for Artificial Intelligence and Decision Support; Center for Medical Statistics, Informatics, and Intelligent Systems; Medical University of Vienna, Austria\\
\email{asan.agibetov@meduniwien.ac.at}
}

\maketitle              

\begin{abstract}
In this work we address the problem of fast and scalable learning of neuro-symbolic representations for general biological knowledge. Based on a recently published comprehensive biological knowledge graph (Alshahrani, 2017) that was used for demonstrating neuro-symbolic representation learning, we show how to train fast (under 1 minute) log-linear neural embeddings of the entities. We utilize these representations as inputs for machine learning classifiers to enable important tasks such as biological link prediction. Classifiers are trained by concatenating learned entity embeddings to represent entity relations, and training classifiers on the concatenated embeddings to discern true relations from automatically generated negative examples. Our simple embedding methodology greatly improves on classification error compared to previously published state-of-the-art results, yielding a maximum increase of $+0.28$ F-measure and $+0.22$ ROC AUC scores for the most difficult biological link prediction problem. Finally, our embedding approach is orders of magnitude faster to train ($\leq$ 1 minute vs. hours), much more economical in terms of embedding dimensions ($d=50$ vs. $d=512$), and naturally encodes the directionality of the asymmetric biological relations, that can be controlled by the order with which we concatenate the embeddings.

\end{abstract}
\section{Introduction}
Over the last decade there has been a very popular trend of merging neural and symbolic representations of knowledge for the large, general-purpose knowledge graphs such as FreeBase~\cite{bollacker_2008} and WordNet~\cite{miller_1995}. The utilized methods can be roughly divided into two groups: i) multi-relational knowledge graph embeddings~\cite{bordes_2013,nickel_2016} and ii) graph embeddings~\cite{perozzi_2014,grover_2016}. The former aims at learning representations of both entities and relations, while the latter focus on the untyped graphs, where each relation's type can be dropped without introducing ambiguities. Both approaches aim at solving the problem of \emph{link prediction}, i.e., modeling the probability of an instance of a relation (e.g., $(u, v) \in r$) based on $d$-dimensional vector representations (e.g., $e(u), e(v), e(r) \in \mathbb{R}^d$) and binary operations defined on them. Thus, in the case of multi-relational knowledge graphs we seek to embed both entities and relations into $d$-dimensional vector space, and we model the probability of a \emph{triple} (labeled arc of a graph) $(u, r, v)$ as $P((u, v) \in r) = \langle e(u) + e(r), e(v) \rangle$ (Euclidean dot product). In the case of unlabeled graphs we drop the labels of the arcs (or edges in case the relations can be treated as symmetric), we therefore do not embed the relations, and model one single arc (or edge) directly as $P(u, v) = \langle e(u), e(v) \rangle$. The Euclidean dot product is only of the many ways to model a probability of having a link (with a label $r$ in the multi-relational case) between the two entities $u, v$. In fact, the underlying geometry may not necessarily be Euclidean, for more in-depth survey of link prediction methodologies please see~\cite{nickel_2016}. In the context of Semantic Web technologies and the Resource Description Framework (RDF) and Web Ontology (OWL) technology stack specialized knowledge graph embedding methodologies have also recently been proposed~\cite{ristoski_2016,cochez_2017}.

In the bioinformatics domain Alshahrani et al.~\cite{alshahrani_2017} recently proposed a novel methodology for representing nodes and relations from structured biological knowledge that operates directly on Linked Data resources, leverages ontologies, and yields neuro-symbolic representations amenable for down-stream use in machine learning algorithms. The authors base their methodology on the DeepWalk algorithm, which performs random walks on the unlabeled and undirected graphs (i.e., with symmetric relations) ~\cite{perozzi_2014} and embeds entities through an approach inspired by the popular Word2Vec algorithm ~\cite{mikolov_2013}.  This methodology is further tuned for multi-relational data by explicitly encoding the sequences of intermingled entities and relations. Such complex intermingled sequences alleviate the innate undirected nature of the random walks, at the expense of increased number of parameters to train. Unfortunately, training such models is computationally expensive (hours on a modern intel core i7 desktop machine) and requires relatively large embedding dimensions ($d = 512$). This manuscript builds upon this seminal work and proposes a more economical, fast and scalable way of learning neuro-symbolic representations. The neural embeddings obtained with our approach outperform published state-of-the-art results, with specific assumptions on the structure of the original knowledge graph, and with the smart encoding of links based on the embeddings of the entities. Among other things the contributions of this work are based on the following hypotheses:

\begin{itemize}
\item There is no need for a sophisticated labeled DeepWalk~\cite{perozzi_2014,alshahrani_2017} to account for all the complexity of the interconnectivity of biological knowledge, since all (considered) biological relations have clear non-overlapping domain and range separations, 
\item We can train faster and more economical log-linear neural embeddings with StarSpace~\cite{wu_2017}, whose quality is comparable to the state-of-the-art results (improves on all but one link prediction task) when considering standard classifiers based on logistic regression as in~\cite{alshahrani_2017},
\item Using the concatenation of the neural embeddings naturally encodes the directionality of the asymmetric biological relations, and fully exploits the non-linear patterns that can be uncovered by the neural network classifiers.
%
%
\end{itemize}
\section{Materials and methods \label{sec:methods}}
\subsection{Dataset and evaluation methodology for link prediction used}

In this work we consider the curated biological knowledge graph, presented in~\cite{alshahrani_2017}. This knowledge graph is based on the three ontologies: Gene Ontology~\cite{ashburner_2000}, Human Phenotype Ontology~\cite{khler_2014} and the Disease Ontology~\cite{kibbe_2015}. It also incorporates the knowledge from several biological databases, including human proteins interactions, human chemical-protein interactions and drug side effects and drug indications pairs. We refer the reader to~\cite{alshahrani_2017} for the detailed description on provenance of the data, and on data processing pipelines employed to obtain the final graph. For the purpose of this work, we summarize the number of biological relation instances present in this knowledge graph in Table~\ref{tab:kg-graph-stats}.

\begin{table}
\centering 
\begin{tabular}{lc} 
\toprule 
    relation    &  number of instances \\ 
\midrule 
    has-target & 554366 \\
    has-disease-annotation & 236259 \\
    has-side-effect & 54806 \\
    has-interaction & 188424 \\
    has-function & 212078 \\
    has-gene-phenotype & 153575 \\
    has-indication & 6704 \\
    has-disease-phenotype & 84508 \\
\bottomrule
\end{tabular} 
\caption{\label{tab:kg-graph-stats} Statistics on the number of edges for the biological relations in the considered knowledge graph~\cite{alshahrani_2017}}
\end{table} 

Our goal is to train fast neural embeddings of the nodes of this knowledge graph, such that we could use these embeddings to perform link prediction. That is, we try to estimate the probability that an edge with label $l$ (e.g., $l=\text{has-function}$) exists between the nodes $v1, v2$ (e.g., $v_1=\text{TRIM28 gene}$ and $v_2 = \text{negative regulation of transcription by RNA polymerase II}$) given their vector representations $\gamma(v_1), \gamma(v_2)$. As in~\cite{alshahrani_2017} we build separate binary prediction models for each relation in the knowledge graph. Note that, in this work we only focus on the link prediction problem where the embeddings are trained on the knowledge graph, in which we remove the 20\% of the edges for a given relation (this corresponds to the first link prediction problem reported in~\cite{alshahrani_2017}). We then use these embeddings to train classifiers (logistic regression and multi-layer perceptron (MLP)) on 80\% of the positive true edges (i.e., relation instances) and on the same amount of generated negative edges. These classifiers are then tested on the remaining 20\% positive and generated negative edges (which have not been used in the embeddings generation). For a fair comparison with the state-of-the-art results, we use the same methodology for negative sample generation, and we use 5-fold cross validation for the training of embeddings and subsequent link prediction classifiers, precisely the same way as in~\cite{alshahrani_2017}. For all of our experiments we do not use any deductive inference, and compare our obtained results with the results obtained without inference in~\cite{alshahrani_2017}.

\subsection{Assumptions on the structure of the Knowledge Graph}
Our methodology exploits the fact that the full biomedical knowledge graph $KG$ we are using only contains relations that can be inferred from the types of the entities that are object and subject of the relation. This means that arc labels can be safely dropped without the loss of semantics and without the introduction of ambiguous duplicated pairs of nodes ($\not \exists r_j . (u, r_j, v) \in KG, r_j \neq r_i \text{ and } (u, r_i, v) \in KG$). Therefore, we can flatten our graph without the risk of having more than one relation connecting the same source and target nodes, i.e., we can simply consider our knowledge graph as a set of pairs of nodes $(u, v)$. As opposed to \emph{DeepWalk} employed by ~\cite{alshahrani_2017}, our methodology does not rely on random walks on knowledge graphs~\cite{perozzi_2014}; instead of producing sequences of labeled entities (nodes and arc labels mixed together), we directly consider pairs of connected nodes. 
Furthermore, we simplify the structure of the knowledge graph by removing anonymous instances that were introduced by the creator of the knowledge graph to assert relation instances in the ABox, i.e., we directly connect OWL classes to de-clutter the graph used to train embeddings. In the original knowledge graph, Alshahrani et al.~\cite{alshahrani_2017} commit to strict OWL semantics when modeling biological relations by asserting anonymous instances, for example a relation instance of \texttt{has-function} (domain: \texttt{Gene/Protein}, range: \texttt{Function}) would be encoded as in Listing~\ref{lst:owl-commitment}, where we present a specific instance of a relation that asserts that the \texttt{TRIM28 gene} has the function of negative regulation of transcription by RNA polymerase II.

\begin{lstlisting}[caption={Biological knowledge representation with OWL semantics commitment},label={lst:owl-commitment},captionpos=b,basicstyle=\scriptsize\ttfamily]
gene: www.ncbi.nlm.nih.gov/gene/
obo: purl.obolibrary.org/obo/
go: aber-owl.net/go/
rdf: www.w3.org/1999/02/22-rdf-syntax-ns#

gene:10155 obo:RO_0000085  go:instance_106358> .
aber-owl:go/instance_106358 rdf:type obo:GO_0000122  .
\end{lstlisting}

\noindent We simplify the knowledge graph by removing all anonymous instances of type {\scriptsize\texttt{<http://aber-owl.net/go/instance\_106358>}} and connecting  entities directly through object relations, i.e., we rewrite all triples of the form presented above (Listing~\ref{lst:owl-commitment}) to the form that only contains object property assertions as demonstrated below (Listing~\ref{lst:owl-commitment-relaxed}).

\begin{lstlisting}[caption={Relaxed biological knowledge representation without OWL semantics commitment},label={lst:owl-commitment-relaxed},captionpos=b,basicstyle=\scriptsize\ttfamily]
gene: www.ncbi.nlm.nih.gov/gene/
obo: purl.obolibrary.org/obo/

gene:10155 obo:RO_0000085 obo:GO_0000122 .
\end{lstlisting}

\noindent We admit such a relaxation in the OWL semantics commitment of the knowledge graph, because we do not leverage any OWL reasoning for our tasks. This relaxation does not change the statistics of the number of biological relation instances present in the knowledge graph (Table~\ref{tab:kg-graph-stats}).

\subsection{Training fast log-linear embeddings with StarSpace}
As opposed to the approach taken by Alshahrani et al~\cite{alshahrani_2017} we employ another neural embedding method which requires fewer parameters and is much faster to train. Specifically, we exploit the fact that the biological relations have well defined non-overlapping domain and ranges, and therefore the whole knowledge graph can be treated as an untyped directed graph, where there is no ambiguity in the semantics of any relation. To this end, we employ the neural embedding model from the StarSpace toolkit~\cite{wu_2017}, which aims at learning \emph{entities}, each of which is described by a set of discrete \emph{features} (bag-of-features) coming from a fixed-length dictionary. The model is trained by assigning a $d$-dimensional vector to each of the discrete features in the set that we want to embed directly. Ultimately, the look-up matrix (the matrix of embeddings - latent vectors) is learned by minimizing the following loss function
$$
\sum_{(a, b) \in E^+, b^- \in E^-} L^{batch} (sim(a, b), sim(a, b_1^-), \ldots, sim(a, b_k^-)).
$$
In this loss function, we need to indicate the generator of positive entry pairs $(a, b) \in E^+$ --  in our setting those are entities $(u, v)$ connected via a relation $r$ -- and the generator of negative entities $b_i^- \in E^-$, similar to the $k$-negative sampling strategy proposed by Mikolov et al.~\cite{mikolov_2013}. In our setting, the negative pairs $(u, v^-)$ are the so-called \emph{negative examples}, i.e., pairs of entities $(u, v^-)$ that do not appear in the knowledge graph. The similarity function $sim$ is task-dependent and should operate on $d$-dimensional vector representations of the entities, in our case we use the standard Euclidean dot product. Please note that the aforementioned embedding scheme is different from a multi-relational knowledge graph embedding task. The main difference is that we do not require the embeddings for the relations.

\begin{figure}[h]
\includegraphics[width=\textwidth]{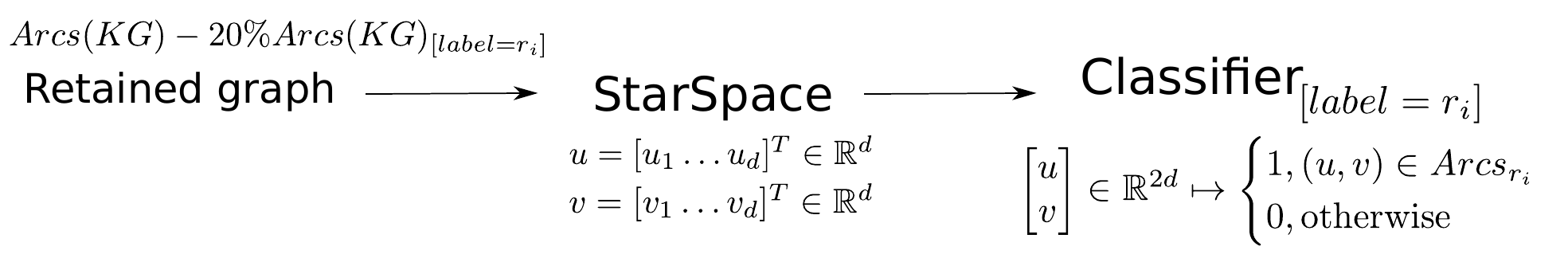}
\caption{\label{fig:concatenated-embedding} Each relation instance $(u, v) \in r$ is represented as a concatenated $[u \mbox{ } v]^T$ vector that preserves \emph{directionality} of the relation $r$, i.e., $(u, v) \in r \neq (v, u) \in r$, $[u \mbox{ } v]^T \in r \neq [v \mbox{ } u]^T \in r$.}
\end{figure}

Based on the embeddings of the nodes of the graph, we can come up with different ways of representing a link between a node $u$ and $v$, as a binary operation defined on the nodes of the graph (see~\cite{grover_2016} for more detail). In particular, we employ the so-called \emph{concatenation} of the embeddings $u, v$ to represent each relation instance as a concatenated vector $[u \mbox{ } v]^T$ (Figure~\ref{fig:concatenated-embedding}).

%
\section{Results}
In Table~\ref{tab:soa-results} we report the state-of-the-art evaluation scores as provided in Alshahrani et al~\cite{alshahrani_2017}. Throughout the rest of this manuscript we refer to these results as \emph{SOTA results} for convenience. We further use these state-of-the-art results to contrast our classification results in Tables~\ref{tab:log-regression} and~\ref{tab:mlp}. To simplify the interpretation of our results, both Tables~\ref{tab:log-regression},~\ref{tab:mlp} report only differences in F-measure and ROC AUC scores for our approach wrt. the SOTA results. Classification results are divided into two parts, differentiated by the classifier used: i) (Table~\ref{tab:log-regression}) logistic regression (as in~\cite{alshahrani_2017}), and ii) (Table~\ref{tab:mlp})) MLP. The two classifiers are trained on \emph{concatenated} embeddings of entities (nodes), which are obtained from the flattened graphs for each biomedical relations via StarSpace~\cite{wu_2017}, as described in Section~\ref{sec:methods}. All classification results presented here are averaged over 5 folds to be directly and fairly compared with the results in~\cite{alshahrani_2017}.

\begin{table}
\centering 
\begin{tabular}{l|cc} 
\toprule 
relation                & F-measure  & ROC AUC \\ 
\midrule 
has-disease-annotation  & 0.89       & 0.95  \\ 
\rowcolor{DarkGray}
has-disease-phenotype   & 0.72       & 0.78  \\
has-function            & 0.85       & 0.95  \\
has-gene-phenotype      & 0.84       & 0.91  \\
\rowcolor{DarkGray}
has-indication          & 0.72       & 0.79  \\
has-interaction         & 0.82       & 0.88  \\
has-side-effect         & 0.86       & 0.93  \\
has-target              & 0.94       & 0.97  \\
\bottomrule
\end{tabular} 
\caption{\label{tab:soa-results} State of the art F-measure and ROC AUC evaluation metrics~\cite{alshahrani_2017}. Rows in dark gray emphasize the worst performing link prediction tasks.}
\end{table}

\subsection{Biomedical link prediction with logistic regression}
Overall, we are able to outperform SOTA results on all relations except for \texttt{has-target} (Table~\ref{tab:log-regression}). It is important to notice that we improve significantly on \texttt{has-indication} and \texttt{has-disease-phenotype} - the two worst performing relations in Alshahrani et al~\cite{alshahrani_2017}. We specifically consider the embeddings of rather small sizes ($[5, 10, 20, 50]$) to emphasize the rapidity and scalability of training embeddings using log-linear neural embedding approaches~\cite{wu_2017}. For all embedding dimensions we train our embeddings for at most 10 epochs, which keeps overall training time of embeddings for one specific biomedical relation under 1 minute on a Core i7 desktop with 32GB of RAM. It is also important to notice that the SOTA results were obtained via the extended \emph{DeepWalk} algorithm~\cite{alshahrani_2017}  with 512 dimensions for the embeddings, which takes several hours to train on our machine. Moreover, our learned embeddings are more consistent, as they have a 0.92-0.99 F-measure and ROC AUC range for all relations, whereas SOTA embeddings range from 0.72 to 0.94.

\begin{table}
\centering
\begin{tabular}{l | cacc | ccca}
\toprule
& \multicolumn{4}{c}{F-measure} & \multicolumn{4}{c}{ROC AUC} \\
\cmidrule(l){2-9}
 &  5 &  10 &  20 &  50 &  5 &  10 &  20 &  50 \\
\midrule
has-disease-annotation &                    -0.027 &                     +0.013 &                     +0.033 &                     +0.071 &                  -0.088 &                   -0.047 &                   -0.028 &                   +0.012 \\
\rowcolor{DarkGray}
has-disease-phenotype  &                    +0.239 &                     +0.260 &                     +0.274 &                     +0.279 &                  +0.180 &                   +0.200 &                   +0.214 &                   +0.219 \\
has-function           &                    +0.013 &                     +0.028 &                     +0.067 &                     +0.117 &                  -0.077 &                   -0.066 &                   -0.030 &                   +0.017 \\
has-gene-phenotype     &                    +0.148 &                     +0.156 &                     +0.159 &                     +0.159 &                  +0.078 &                   +0.086 &                   +0.089 &                   +0.089 \\
\rowcolor{DarkGray}
has-indication         &                    +0.186 &                     +0.262 &                     +0.270 &                     +0.275 &                  +0.112 &                   +0.192 &                   +0.200 &                   +0.205 \\
has-interaction        &                    +0.010 &                     +0.147 &                     +0.179 &                     +0.180 &                  -0.034 &                   +0.088 &                   +0.119 &                   +0.120 \\
has-side-effect        &                    +0.091 &                     +0.105 &                     +0.128 &                     +0.137 &                  +0.021 &                   +0.036 &                   +0.059 &                   +0.067 \\
has-target             &                    -0.107 &                     -0.077 &                     -0.047 &                     -0.018 &                  -0.109 &                   -0.083 &                   -0.057 &                   -0.034 \\
\bottomrule
\end{tabular}
\caption{\label{tab:log-regression} Differences in F-measure and ROC AUC scores for our classification results for logistic regression models trained on our neural embeddings wrt. the SOTA results. In light gray are the minimal embedding dimension with the better scores than the state of the art (excluding \texttt{has-target} relation). Rows colored with dark gray represent the worst performing SOTA relations, which we outperform significantly.}
\end{table}

\subsection{MLP and  biomedical link prediction}

We hypothesize that our approach of augmented embedding dimension via concatenation of entity embeddings is more suited for neural network architectures. Indeed, we are able to obtain very good biological link prediction classifiers by using concatenated embeddings and multi-layer perceptrons. We experimeted with different shallow and deep architectures (hidden layer sizes ([200], [20, 20, 20], [200, 200, 200]), which yielded almost similar performances.  The results of a shallow neural networks with one hidden layer consisting of 200 neurons are summarized in Table~\ref{tab:mlp}, that empirically show that the concatenation of the neural embeddings to represent a link between the two entities fully exploits the non-linearity patterns, which can be uncovered by the neural network classifiers. As a result, we are able to improve the SOTA results for all the biological link prediction tasks.
%
%
%
%

\begin{table}
\centering
\begin{tabular}{l | cacc | ccca}
\toprule
& \multicolumn{4}{c}{F-measure} & \multicolumn{4}{c}{ROC AUC} \\
\cmidrule(l){2-9}
&  5 &  10 &  20 &  50 &  5 &  10 &  20 &  50 \\
\midrule
has-disease-annotation &                    +0.095 &                     +0.109 &                     +0.110 &                     +0.110 &                  +0.035 &                   +0.049 &                   +0.050 &                   +0.050 \\
\rowcolor{DarkGray}
has-disease-phenotype  &                    +0.272 &                     +0.279 &                     +0.280 &                     +0.280 &                  +0.212 &                   +0.219 &                   +0.220 &                   +0.220 \\
has-function           &                    +0.148 &                     +0.150 &                     +0.149 &                     +0.150 &                  +0.048 &                   +0.050 &                   +0.049 &                   +0.050 \\
has-gene-phenotype     &                    +0.160 &                     +0.160 &                     +0.160 &                     +0.160 &                  +0.089 &                   +0.090 &                   +0.090 &                   +0.090 \\
\rowcolor{DarkGray}
has-indication         &                    +0.276 &                     +0.278 &                     +0.279 &                     +0.279 &                  +0.206 &                   +0.208 &                   +0.209 &                   +0.209 \\
has-interaction        &                    +0.180 &                     +0.180 &                     +0.180 &                     +0.180 &                  +0.120 &                   +0.120 &                   +0.120 &                   +0.120 \\
has-side-effect        &                    +0.128 &                     +0.137 &                     +0.139 &                     +0.139 &                  +0.058 &                   +0.067 &                   +0.069 &                   +0.069 \\
has-target             &                    -0.024 &                     +0.006 &                     +0.023 &                     +0.033 &                  -0.040 &                   -0.016 &                   -0.003 &                   +0.006 \\
\bottomrule
\end{tabular}
\caption{\label{tab:mlp} Differences in F-measure and ROC AUC scores for our classification results with MLP models with one hidden layer consisting of 200 hidden units, trained on our neural embeddings wrt. the SOTA results. In light gray are the minimal embedding dimension with the better scores for \textbf{all} relations than the state of the art. Rows colored with dark gray represent relations where the previous SOTA approach performs worst and where our approach outperforms significantly.}
\end{table}

\section{Discussion and conclusion}
Recent trends of neuro-symbolic embeddings continue the long-sought quest of the artificial intelligence community to unify the two disparate worlds, where the reasoning is performed either in a discrete symbolic space or in a continuous vector space. As a community, we are still somewhere along this road, and up to date there has still been no evidence of a clear way of combining the two approaches. The neuro-symbolic representations based on random walks on RDF data for the general biological knowledge as introduced by~\cite{alshahrani_2017} are an important first development. The methodology allows for leveraging the existing curated and structured biological knowledge (Linked Data), incorporating OWL reasoning, and enabling the inference of hidden links that are implicitly encoded in the biological knowledge graphs. However, as our results demonstrate, it is possible to obtain improved classification results for link prediction if we relax the constraints of multi-relational biological knowledge structure, and consider all arcs as part of one semantic relation. Such a relaxation gives rise to faster and more economical generation of neural embeddings, which can be further used in scalable downstream machine learning tasks. While our results demonstrate excellent prediction performance (all F-measure and ROC AUC scores range in 0.92-0.99), they outline that having very well-structured input data is a core ingredient. Indeed, the biological knowledge graph curated by Alshahrani et al.~\cite{alshahrani_2017} implicitly encodes significant biological knowledge available to the community, and simple log-linear embeddings coupled with shallow neural networks are enough to obtain very good prediction results for the transductive link prediction problems. Unfortunately, the quest of merging symbolic and continuous representations cannot be fulfilled to its advertised limits, as was already mentioned in~\cite{alshahrani_2017}, symbolic inference (OWL-EL reasoning) do not yield significant improvements on link prediction tasks. Indeed, we managed to get very good scores without any deductive completion of the Abox of the knowledge graph. Another important aspect which we implicitly emphasized in our work is the evaluation strategy of the neural embeddings. When dealing with big and rich knowledge graphs one has to meticulously generate train and test splits, which avoid potential leakage of information between the two sets. Failing to do so might lead to the models which overfit and are unable to truly perform link predictions. As part of our future work we would like to focus on the creation of different evaluation strategies that test the quality of the neural embeddings, their explainability, and we would like to consider not only transductive link prediction problems, but also focus on the more challenging inductive cases.

%
%

\end{document}